\documentclass[journal]{IEEEtran}
\ifCLASSINFOpdf
  \usepackage[pdftex]{graphicx}
\else
\fi
\usepackage{rotating}
\usepackage{makecell}
\usepackage{multirow}
\usepackage{colortbl}
\usepackage{xcolor}

\usepackage{etoolbox} 
\usepackage{tikz}
\newcommand\copyrightnotice[1]{
    \begin{tikzpicture}[remember picture,overlay]
    \node[anchor=south,yshift=12pt] at (current page.south) {\fbox{\parbox{\dimexpr\textwidth-\fboxsep-\fboxrule\relax}{#1}}};
    \end{tikzpicture}
}

\usepackage{amsmath}

\begin{document}

\bstctlcite{IEEEexample:BSTcontrol} 
%
\title{Task Dynamics of Prior Training Influence Visual Force Estimation Ability During Teleoperation}
%
%
%

\author{Zonghe Chua,~\IEEEmembership{Student Member,~IEEE,}
        Anthony M. Jarc,
        Sherry Wren,\\
        Ilana Nisky,~\IEEEmembership{Senior Member, IEEE,}
        and Allison M. Okamura,~\IEEEmembership{Fellow, IEEE}

\thanks{Manuscript received .....}%
\thanks{This work was supported by Stanford University, Intuitive Surgical, Inc., the Israeli Science Foundation (Grant 823/15), the Israeli Ministry of Science and Technology via the Israel-Italy Virtual Lab on Artificial Somatosensation for Humans and Humanoids, and the Helmsley Charitable Trust through the Agricultural, Biological \& Cognitive Robotics Initiative of Ben-Gurion University of Negev.}%
\thanks{Z. Chua and A. M. Okamura are with the Department of Mechanical Engineering, Stanford University, Stanford, CA 94305, USA}
\thanks{S. Wren is with the Department of Surgery, Stanford University and Palo Alto Veterans Affairs Health Care System, Palo Alto, CA 94304 USA}
\thanks{A. M. Jarc is with Intuitive Surgical, Inc., Sunnyvale, CA 94086, USA}
\thanks{I. Nisky is with the Department of Biomedical Engineering and with the Zlotowski Center for Neuroscience, Ben-Gurion University of the Negev, Beer-Sheva 84105, Israel }

}

%
%

\markboth{ }%
{Shell \MakeLowercase{\textit{et al.}}: Bare Demo of IEEEtran.cls for IEEE Journals}
%



\maketitle

\copyrightnotice{\scriptsize \copyright \, 2020 IEEE. Personal use of this material is permitted. Permission from IEEE must be obtained for all other uses, in any current or future media, including reprinting/republishing this material for advertising or promotional purposes,creating new collective works, for resale or redistribution to servers or lists, or reuse of any copyrighted component of this work in other works.}

\begin{abstract}
The lack of haptic feedback in teleoperation is a potential barrier to safe handling of soft materials, yet in Robot-assisted Minimally Invasive Surgery (RMIS), haptic feedback is often unavailable. Due to its availability in open and laparoscopic surgery, surgeons with such experience potentially possess learned models of tissue stiffness that might promote good force estimation abilities during RMIS. To test if prior haptic experience leads to improved force estimation ability in teleoperation, 33 naive participants were assigned to one of three training conditions: manual manipulation, teleoperation with force feedback, or teleoperation without force feedback, and learned to tension a silicone sample to a set of forces. They were then asked to perform the tension task, and a previously unencountered palpation task, to a different set of forces under teleoperation without force feedback. Compared to the teleoperation groups, the manual group had higher force error in the tension task outside the range of forces they had trained on, but showed better speed-accuracy functions in the palpation task at low force levels. This suggests that the dynamics of the training modality affect force estimation ability during teleoperation, with the prior haptic experience accessible if formed under the same dynamics as the task.

\end{abstract}


%
\IEEEpeerreviewmaketitle

\section{Introduction}
%
%
%
%

\IEEEPARstart{C}{urrent} instruction in robot-assisted minimally-invasive surgery (RMIS) is not necessarily delivered in a standardized way \cite{Fisher2015}. Trainees come into programs with varying amounts of experience in open and laparoscopic surgical skills \cite{GobernSurvey}\cite{Knab2018}. Laparoscopic surgery has many similarities with RMIS, such as the viewing of the surgical scene through a camera, the use of surgical manipulators, and the scaling of input to output movements. However, in laparoscopy, the direction of movement at the handle is opposite that of the tooltip due to the tool's pivoting about its insertion point. The resulting fulcrum effect has been shown to scale movements and forces, which combine to affect perceived environment stiffness \cite{Nisky2012}. Unlike RMIS and laparoscopy, open surgery does not rely on surgical manipulators and lacks any form of motion scaling. The similarities and differences between these traditional forms of surgery (open and laparoscopic) and RMIS, have motivated the study of whether prior experience in the traditional forms of surgery can help shorten the steep learning curve of RMIS \cite{Mazzon2017} and whether such training should be a pre- or co-requisite of an RMIS curriculum.

A majority of the works that have attempted to measure skill transfer to RMIS have focused on transfer from laparoscopy, with overall skill quantified by a proprietary global skill metric, the MScore \cite{Teishima2012}\cite{Yoo2015}\cite{Finnerty2016}\cite{Pimentel2018}. The MScore is implemented on the da Vinci Skills Simulator software (Mimic Technologies Inc., Seattle, WA, USA), and has been shown to have construct validity \cite{finnegan2012vinci}\cite{alzahrani2013validation}\cite{liss2012validation}. Previous work suggests that the transferability of skill to RMIS is limited, with majority of the studies showing no significant differences between laparoscopically experienced surgeons and novices in simulator tasks for RMIS \cite{Teishima2012}\cite{Yoo2015}\cite{Pimentel2018}. In contrast, Finnerty et al. found that laparoscopic experience was associated with significantly higher MScores in the tasks they measured \cite{Finnerty2016}. One limitation of these previous works is that the MScore is dominated by dexterity metrics \cite{hung2011face}, with only one component metric, duration of excessive instrument force, quantifying some aspect of force modulation. Furthermore, it is unclear how accurately the physics and instrument forces are modeled in the da Vinci Skills Simulator. 
Thus, these prior works that rely on the MScore offer limited insight into whether laparoscopic or open surgical skill can aid in force modulation and estimation in RMIS.

Effective tissue handling is a key component of surgical skill that is included in surgical assessment rubrics like the Objective Structured Assessment of Technical Skills (OSATs) \cite{martin1997osats} and the Global Evaluative Assessment of Robotic Skills (GEARs) \cite{goh2012global}. Effectively handling tissue in RMIS is thought to be difficult because of the lack of haptic feedback, which has been shown to be useful in force modulation in RMIS. Prior work has found that including haptic feedback in RMIS reduces force application in a variety of surgical exercises such as blunt dissection \cite{Wagner2007}, tissue grasping \cite{King2009}, and puncturing \cite{Talasaz2017}. However, the lack of cost effective, biocompatible, and sterilizable force sensors has prevented commercially feasible implementation of haptic feedback. Thus in RMIS, the ability to estimate force without haptic feedback is of high importance relative to other applications of teleoperation.

In current RMIS, the lack of haptic feedback requires that surgeons rely on vision to perform force estimation. Visual cues about force in RMIS include visible tissue damage and the blanching of tissue due to restricted blood flow when it is stretched or squeezed. In visual estimation of force, the close association of force to visual cues during object interaction allows for the development of visual mappings to force that are largely heuristic \cite{Klatzky2014}. The ability to use such learned mappings to estimate force visually has been demonstrated previously in a tape-pulling task \cite{Michaels1998}.

A more general mapping is that of stiffness, which requires developing a model relating displacement to force. Stiffness is a higher-order material property that potentially allows for generalized force estimation outside of learned movements and scenarios. The perception and estimation of stiffness have been shown to be formed under multimodal cue integration of displacement and force percepts, in which haptics, when present, play a dominant role in force perception relative to vision \cite{Kuschel2010}\cite{Cellini2013}.

In RMIS, forces on tissue can be estimated visually as mentioned previously, while displacement on tissue can be sensed both visually and through proprioception. However, in open and laparoscopic surgery, the development of estimates of tissue stiffness is informed by the presence of haptic feedback. The coupling of visual cues to kinesthetic force would then allow for the development of a prior expectation of what a certain amount of perceived tissue displacement ``feels like". A similar phenomenon resulting from strong visuohaptic correlations is that of the limb embodiment described in the rubber hand illusion \cite{botvinick1998rubber}\cite{Samad2015}. According to Kording and Wolpert, these prior expectations can inform our movement planning and sensorimotor control \cite{KordingWolpert2006}. Thus, if these haptically informed stiffness models were accessible during RMIS, then surgeons with prior experience in open or laparoscopic surgery would potentially have better force estimation abilities compared to those with no experience.

\begin{figure}[!b]
\centering
\includegraphics[width=\linewidth]{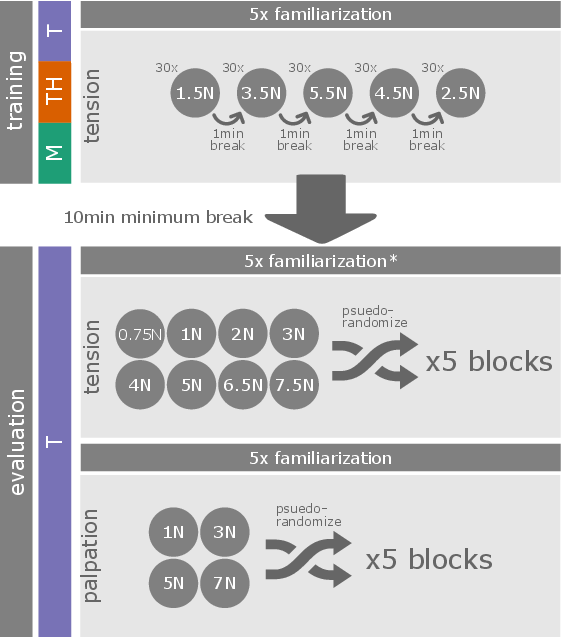}
\caption{Protocol for the experiment. Participants are assigned to one of three conditions during training: manual manipulation (M), teleoperation with haptics (TH), or teleoperation without haptics (T). They are then evaluated under the T condition in the tension and palpation tasks. The reference force levels are presented in pseudo-randomized blocks. \protect\\ * The familiarization before the evaluation tension task is only performed if the participant was assigned to the M condition.}
\label{exp_protocol}
\end{figure}

\begin{figure*}[!t]
\centering
\includegraphics[width=\textwidth]{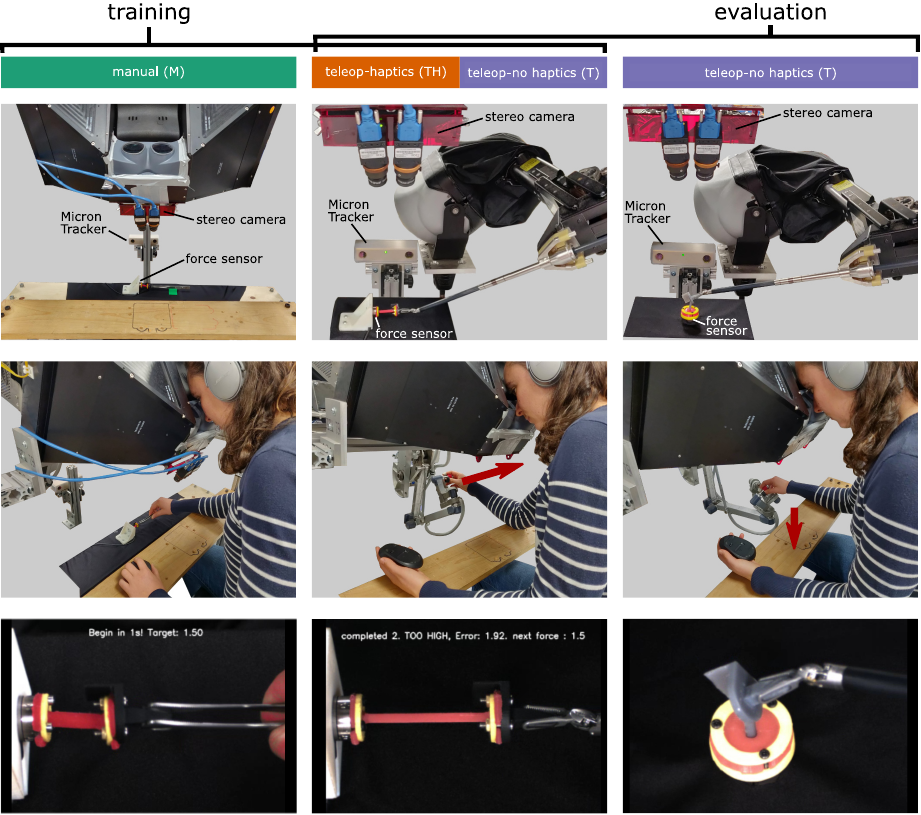}
\caption{Hardware setups for different haptic training conditions in the training block and the evaluation block. Rows: (top) Setup for the manipulation of the silicone sample, (middle) how the participant interacts with the experimental hardware and their direction of pull (red arrow), and (bottom) view of the task environment through the surgeon console. Columns: (left) The tension task for the manual condition, showing the target force prompt provided at the beginning of each trial during training, (middle) the tension task for the teleoperated condition with the text feedback at the end of a trial during training, and  (right) the palpation task for the teleoperated condition during evaluation.}
\label{exp_setup}
\end{figure*}

To date, there has been little work exploring how prior experiences with haptic feedback influence visual force estimation ability during teleoperation. Thus in this work, we investigated the role of prior haptic experience, a feature of both open and laparoscopic surgical training, on force estimation in teleoperation with only visual feedback. While surgeons exert forces and moments in different directions and from shifting viewpoints during surgery, the skill can take a long time to acquire \cite{Mazzon2017}, and experience can vary among groups of similar background \cite{Teishima2012}\cite{Pimentel2018}. Furthermore, such studies using surgeons as subjects can suffer from small numbers of subjects \cite{Teishima2012}\cite{Yoo2015}\cite{Pimentel2018}. To ensure timely development of a consistent prior stiffness model over an adequately sized subject pool, we abstracted the task of tissue retraction to focus on tension exerted in a single direction, and recruited naive participants. Specifically, we asked them to learn to exert different amounts of force in a material tensioning task under the condition of either manual manipulation, teleoperation with haptic feedback, or teleoperation without haptic feedback. We then evaluated the participants' ability to estimate force in the same task under teleoperation without haptic feedback. Additionally, we evaluated their ability to infer the stiffness estimate of the same material in compression and perform force estimation, through a palpation task under teleoperation without haptic feedback. 

While we hypothesize that participants can exploit the haptic information present during training to develop prior expectations of force that can aid in force estimation performance in the absence of haptic feedback, it is also possible that they become reliant on the haptic feedback to perform the task and suffer a degradation in performance when it is lacking. This high reliance on available haptic feedback for stiffness estimation has been documented by Kuschel et al. \cite{Kuschel2010} and Cellini et al. \cite{Cellini2013}. In the palpation task, the change in the stiffness characteristics of the material, and the difference in the direction of the performed movements, should reduce the direct influence of both visual and haptically informed stiffness estimates learned from the tension task. However, participants from the groups that trained with haptics should still be able to integrate the material stiffness estimate they learned in training with those from their experience interacting with viscoelastic materials in their day-to-day activities, to better estimate their applied forces compared to those that trained without haptics.

\begin{figure}[!t]
\centering
\includegraphics[width=\linewidth]{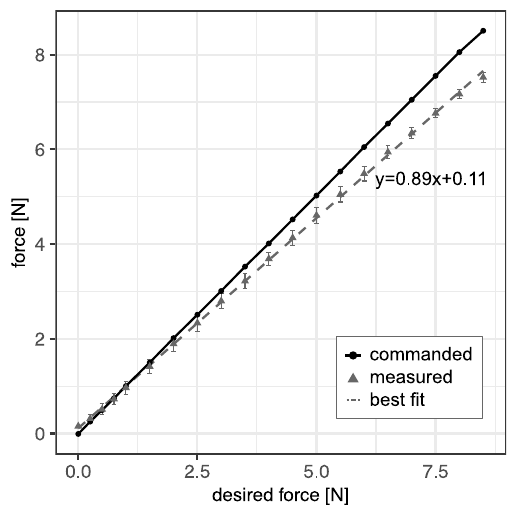}
\caption{Measured force in the direction of pull commanded by the dVRK via joint torques (dots), and measured at the end effector (triangles) versus commanded force. The dashed line is the line of best fit to the measured forces. Error bars represent standard deviation.}
\label{forceverification}
\end{figure}

\section{Methods}

\subsection{Participants}

We recruited 33 participants, between the ages of 18 and 36. All participants were right-handed and were balanced for sex over all haptic training conditions. All participants were novices in that they either had never used a da Vinci Surgical System or da Vinci Research Kit (Intuitive Surgical, Inc., Sunnyvale, CA, USA) \cite{kazanzides2014dvrk} before or had only used one in a brief demonstration setting. Three participants' results were not analyzed due to one being unable to learn how to teleoperate the robot smoothly, one pulling on the sample too hard and permanently damaging it, and one who did not wear their headphones properly during the training block.

\subsection{Experiment Design}
The experiment was a mixed design, with one between-subjects factor and two within-subjects factors. The between-subjects factor was the haptic training condition: Either manual manipulation (M), teleoperation with haptics (TH), or teleoperation without haptics (T). Each participant was assigned to one of the conditions and learned to perform a material tensioning task over a set of reference force levels (Fig.\,\ref{exp_protocol}). They were then evaluated in a block consisting of two sub-experiments. In the first sub-experiment, participants performed the same tensioning task under the T condition over a different set of reference force levels. In the second sub-experiment, participants performed a palpation task under the T condition over a separate set of reference force levels. The within-subjects factors were the blocks of pseudo-randomized trials and the set of reference force levels the participants were asked to exert. The range of reference force levels for both the training and evaluation blocks were within the 0.5-12\,N reported for maneuvers such as cutting, suturing, dissecting, and grasping in minimally invasive surgery \cite{Toledo1999}.

\subsection{Setup}

\subsubsection{Hardware}
There were three hardware setups that were required for this experiment. One was for the tension task under the M condition. Two were for the teleoperated conditions (T and TH), with the first being for the tension task and the other being for the palpation task. All teleoperated conditions used the full da Vinci Research Kit (dVRK), consisting of the surgeon console, master terminal, and patient side robot. Teleoperation control was achieved through unilateral proportional-derivative (PD) control with gains tuned for tracking and stability. For the TH condition, position-force teleoperation was used, such that the commanded force output was equal to the force measured by the force sensor. The M condition used only the surgeon console. Each of the setups is shown in Fig.\,\ref{exp_setup} and described in detail below.

In the tension task for manual manipulation, a red silicone sample with an attached rigid end piece was affixed to a 6-axis Nano17 force-torque sensor (ATI Industrial Automation, Apex, NC, USA) and mounted horizontally to a base fixture. This base fixture assembly was placed beneath the stereoscopic displays of the dVRK, within reach of a user sitting at the console. A black cloth was used to create a monochrome backdrop. Video of the stage was captured by a stationary stereo camera assembly consisting of two Flea3 cameras (FLIR Systems, Wilsonville, OR, USA) and streamed in real-time to the stereoscopic displays. The displacement of the silicone sample was measured by tracking the position of the rigid end piece using the MicronTracker (ClaroNav, Toronto, ON, Canada), a camera-based marker tracking system. A pair of surgical forceps was also provided in this setup for participants to use to grip the end piece.

In the teleoperated \textit{tension task}, the base fixture assembly was located within the workspace of the right patient-side manipulator (PSM) of the dVRK. The stereo camera assembly was positioned in the environment at the same distance away from the fixture as in the manual manipulation condition. The MicronTracker was used for tracking the position of the rigid end piece. Participants under this condition would teleoperate the right PSM through the right master terminal manipulator (MTM) to manipulate the silicone sample. The motion scaling ratio of the master-side to the patient-side was set to 2:1. 

In the TH condition, the axial forces measured by the force sensor were used to provide kinesthetic force feedback to the MTM at a rate of 1\,kHz. The force was commanded to the MTM open loop, so we characterized the relationship between the desired and rendered force in the direction of pull while the MTM was in the configuration adopted at the start of each trial. The ratio of desired force to rendered force was 0.89, and the measured force had a constant offset of 0.11\,N at 0\,N (Fig.\,\ref{forceverification}). The ratio is near the human just noticeable difference limit of 11\% and 13\% reported for kinesthetic force perception during extension of the wrist and elbow respectively, in a similar bent-arm configuration, but with a different haptic device \cite{Feyzabadi2013}. While the offset of 0.11\,N is above the 0.04\,N absolute thresholds \cite{Feyzabadi2013}, this effect is negligible over the duration of the pull. Additionally, any force above 8.5\,N commanded via joint torques was limited by the dVRK software. Thus, our experiment limited the maximum force that participants were allowed to exert on the silicone samples.

In the teleoperated \textit{palpation task}, a cylindrical silicone sample was mounted to the force sensor, and placed within the workspace of the right PSM, with the palpation surface facing up. Participants teleoperated the right PSM to manipulate a rigid probe, which was tracked by the MicronTracker.

In all conditions, blocks, and tasks, participants were provided a wireless clicker which they used to confirm their final estimate of the applied force on the sample at the end of each trial.

\subsubsection{Fabrication and Characterization of Silicone Samples}

Silicone samples were injection molded using DragonSkin 10 (Smooth-On Inc., Macungie, PA, USA) in the manufacturer recommended 1:1 ratio by weight. These silicone samples had two flanges and a 5\,mm diameter stem that was 20\,mm in length. Red coloring was added using Sil-Pig dye (Smooth-On Inc., Macungie, PA, USA). To assess the consistency of the manufactured samples, each silicone sample was mounted to the experimental setup and was pulled by the dVRK PSM to a displacement of 45\,mm. 

\begin{figure}[!t]
\centering
\includegraphics[width=\linewidth]{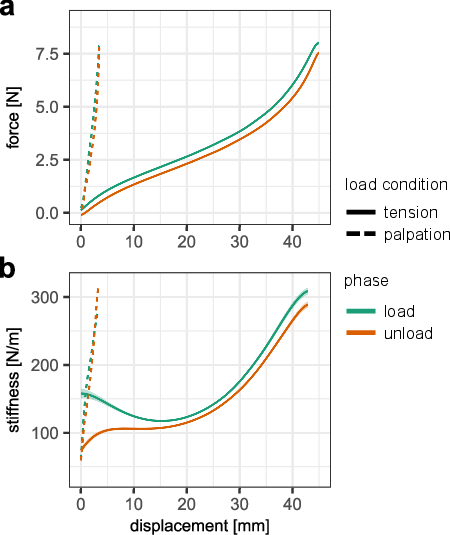}
\caption{(a) The mean force-displacement loading (green) and unloading curves (orange) for the 30 tension samples each pulled twice (solid) and the single palpation sample palpated twice (dotted). (b) The corresponding average stiffness of the samples over displacement. Error ribbons denote 95\% confidence intervals.}
\label{characterization}
\end{figure}

A single red cylindrical silicone sample with a diameter of 30\,mm and a width of 7\,mm was produced for the palpation task using DragonSkin 10. This was also characterized by the dVRK PSM performing palpations up to a force of 7.5\,N. The results of the tests are summarized in Fig.\,\ref{characterization}. 

For the range of forces tested in tension and palpation, the stiffness of the samples ranged from 59-313\,N/m, resulting in Young's moduli in the range of 15-80\,kPa. This is similar to spleen and cardiac muscle tissue at the low end of the range, and bladder and skeletal muscle tissue at the high end \cite{Guimaraes2020}. For both types of characterization, the displacement rate was set to 7.5\,mm/s. 

\subsubsection{Force-based Color Change Rendering}
Color change is thought to be an important visual cue for force in RMIS as blanching of tissue can indicate a large amount of applied force. To present a form of force-based color change feedback to the participants in our study, real-time video processing with the OpenCV library \cite{opencv_library} was used to generate color changes in the silicone sample. Color thresholding was performed on the sample and a red colored semi-opaque overlay was superimposed with a 30\% transparency. To date, no work has been done to quantify the mapping of color changes to tool interactions in tissue. One way to approximate this environmental behavior is to linearly map color saturation of the overlay to the horizontal force measured by the force sensor. This mapping was defined as
\begin{equation}
\label{sat_eqn}
    S = 255\left(1 - \frac{F}{F_{\text{sat}}}\right) \ \text{,}
\end{equation}
where $S$ is the saturation value for the overlay, $F$ is the measured force from the force sensor and $F_{\text{sat}}$ is the predefined saturation force. A value of $S=255$ would correspond to the sample appearing a deep red and $S=0$, a very light pink. In the experiment, $F_{\text{sat}}$ was set to 8.2\,N so that S would be 0 at 8.2\,N. This was below maximum force limit of 8.5\,N that was commandable through the MTM under the TH condition during the tensioning task. If 8.2\,N was exceeded, the overlay was immediately colored brown to indicate to the participants that they had reached an excessive level of force. The same color mapping was used in palpation, with the force in Eq.(\ref{sat_eqn}) being a compressive force instead of a tensile force.

\subsection{Procedure}

\subsubsection{Training Block}

Participants were trained to exert a set of 5 force levels by tensioning the silicone sample under their assigned haptic training condition. The forces to be learned were presented in the order: 1.5\,N, 3.5\,N, 5.5\,N, 4.5\,N  and 2.5\,N. Each force level was practiced 30 times each with a break of 1 minute after each level was completed (Fig.\,\ref{exp_protocol}). Before the start of the training block, participants were given a minimum of 5 familiarization trials to ensure that they could teleoperate the robot, were making the correct pulling movements, and knew how to confirm their guesses.

For each trial, participants were asked to horizontally tension the sample to the point at which they thought the force they were exerting on the sample matched the target reference force. Once they had confirmed their guess using the provided clicker in their left hand, they were presented with the results of their attempt. This feedback consisted of displaying their numerical force error and also qualitative feedback (Fig.\,\ref{exp_setup}, third row). The latter consisted of a classification of “too high”, “too low” or “correct”. The range in which a guess was deemed “correct” was determined during sample characterization prior to conducting the experiment. This was done by fitting a 3rd-order polynomial to the material characterization data points and finding the upper and lower force bounds corresponding to 2\,mm from the target force-displacement point on the curve. This threshold for accuracy was determined empirically from pilot studies to reduce excessive trial-to-trial variability during learning.

To encourage consistency in their movements, all participants were instructed to keep their elbow position fixed, and to perform the horizontal pull motion by pivoting around their elbow. In addition, before the start of each trial, the MTM is reset to a fixed home pose. There was a soft time limit of 7 seconds during each trial. After this time was exceeded, a 1\,kHz tone was displayed to the participant through a pair of headphones. This tone ended only when the participant confirmed their guess for the current trial. Participants were told to attempt their guess before the tone was emitted.


To ensure the integrity of the silicone sample, participants were instructed to stay below the 8.2\,N force threshold, above which the color overlay saturated abruptly to brown. They were not informed of the threshold force value. 

In pilot testing with a fixed viewpoint, we found that participants relied heavily on the boundaries of their field of view as a reference for the tip position. However, such a strategy would not be possible in applications such as RMIS due to the camera movement. Thus, to reduce participants’ reliance on fixed reference points in their field of view, without altering the magnification of the scene, the camera position was horizontally displaced at random in each trial. This was done by cropping the image and then shifting the cropped window horizontally on the original video frames.

\begin{figure}[!t]
\centering
\includegraphics[width=\linewidth]{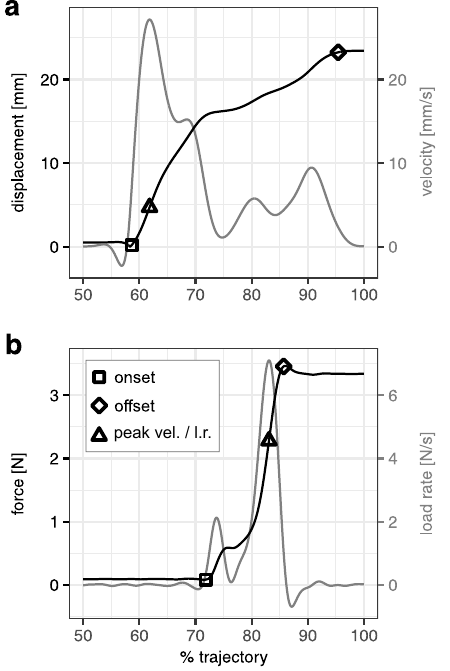}
\caption{(a) The result of applying the criteria for movement onset, offset, and peak velocity to a displacement trajectory for the tension task. (b) The result of applying the criteria for movement onset, offset, and peak load rate to a force trajectory for the palpation task. The force trajectory was used in the palpation task because the position of the probe tip was not a reliable measure of sample displacement.}
\label{performanceMetrics}
\end{figure}

\subsubsection{Evaluation Block}

In the tension sub-experiment, participants were asked to exert a set of 8 forces (0.75\,N, 1\,N, 2\,N, 3\,N, 4\,N, 5\,N, 6.5\,N, 7.5\,N), 5 times each in pseudo-random order (Fig.\,\ref{exp_protocol}). Unlike in the training block, they were not provided with knowledge of results after each trial. Before beginning the trials, participants were once again reminded to stay below the visually indicated force thresholds. Between the training block and the first evaluation task, a minimum 10 minute break was enforced to allow for the experimenter to reconfigure the setup and the participant to rest. Because participants under the M condition had not yet been introduced to the robot, they were given a minimum of 5 familiarization trials to ensure that they knew how to teleoperate the robot. 

In the palpation sub-experiment, participants were instructed to teleoperate the dVRK to grip the probe and palpate the red cylindrical silicone sample vertically (Fig.\,\ref{exp_setup}, third column). Four force levels (1\,N, 3\,N, 5\,N and 7\,N) were presented 5 times each in pseudo-random order (Fig.\,\ref{exp_protocol}). The soft time limit was also maintained at 7 seconds. Participants were instructed to keep the probe as vertical as possible during palpation and to attempt to palpate the center of the sample. Because there was no easy way for participants to use external reference points in the palpation task, the camera position was not randomized as it was in the tension task.

At the end of the evaluation block, participants were asked to fill in a survey to rank the types of stimuli they used the most during each of the blocks and tasks, and to indicate their experience levels with haptic devices, surgical robots, and viscoelastic materials. 

\subsection{Performance Metrics}

Performance metrics were computed for all valid trajectories. A trajectory was considered valid if the pulling or palpating motion did not contain a false start or more than two reversals. The latter indicates that the participant was not attempting to make a single guess but was probing the system for more information. 

In this study, the performance of each participant was quantified by accuracy and speed. Accuracy was measured as force error. This was computed by taking the difference between the force applied to the sample when the user confirmed their guess at the end of their pull, and the target reference force for that attempt. Thus, a positive force error would correspond to overestimation, and vice versa, with values closer to zero implying better accuracy.

Speed was measured using the peak velocity and movement time. In the tension task, the movement onset was computed by identifying the time when velocity first crossed 25\% of the peak initial velocity (Fig.\,\ref{performanceMetrics}a). The movement offset was found by finding the last time that the velocity passed 25\% of the peak final velocity. The difference between the movement offset and onset times then defined the movement time. Because the end cap that the participant pulled on was attached directly to the sample endpoint, the position information from the MicronTracker was usable in this task. For the palpation task, the probe was not attached to the sample, and thus the amount of displacement of the sample by the probe tip did not correspond with the position data from the MicronTracker. Thus, in order to compute the movement time, the force profile was used instead of the displacement profile, and peak load rates were used instead of peak velocity (Fig.\,\ref{performanceMetrics}b).  

As shown by Fitts, slower movements are more accurate \cite{fitts1954information}. The speed-accuracy tradeoff can be quantified as a function of movement time and error \cite{plamondon1997speed}\cite{Nisky2014}\cite{Coad2017}. A better speed-accuracy function implies that better accuracy can be achieved at equal or faster task performance speed relative to another speed-accuracy function. We chose here to account for the difference in units of accuracy and time by defining a normalized metric, normalized absolute error-time (NAET), as a product of movement time and absolute force error, such that 
\begin{equation}
\text{NAET} = \frac{|\text{error}|}{|\text{error}_{\text{max}}|} \times \frac{\text{time}}{\text{time}_{\text{max}}} \ \text{,}
\end{equation}
where $\text{error}_{\text{max}}$ and $\text{time}_{\text{max}}$ are the observed maxima for each quantity during evaluation over all subjects (i.e., all three training conditions). The absolute value of force error is used because taking the product of force error with time reduces the interpretability of the direction component of the metric. For instance, two trials with similar movement times and force error magnitudes, but with opposing force error signs, should be more similar to each other in terms of speed-accuracy than two trials with different movement times and force errors that are similar in both magnitude and sign. The normalization of error and time over their population maximums give unitless quantities between 0 and 1 that assign equal weight to error and time. A lower NAET implies that a participant makes less trade-off between speed and accuracy compared to another participant with higher NAET.

\begin{figure}[!t]
\centering
\includegraphics[width=0.95\linewidth]{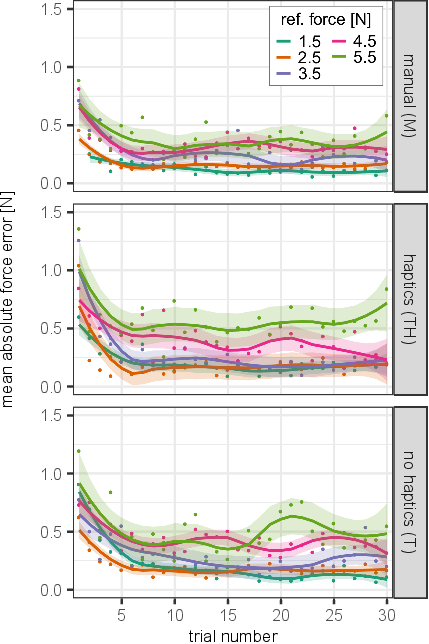}
\caption{Learning curves for each reference force level over training 
conditions. Points represent mean force error over all participants for that condition and force level during each trial. Locally estimated scatterplot smoothing curves are fit to each reference force. Error ribbons denote 95\% confidence intervals.}
\label{Learning curves} 
\end{figure}

\subsection{Statistical Analysis}

A linear mixed-effects model was fit for each performance metric. The fixed effects were haptic training condition, reference force, experiment progression (trial number), and all their interactions. The random effect was subject modeled as a random intercept. Metrics that were positive valued were log-transformed to fit the assumptions for normality required by the model. An F-test using Satterthwaite's approximation for the denominator degrees of freedom was performed for each metric to test for significance. For each fixed effect, a p-value of less than 0.05 was deemed as significant. If a fixed effect or an interaction with the haptic training condition was found to be significant, a post-hoc t-test was performed using the Bonferroni correction to test for pairwise differences between conditions.  

\subsection{Estimated Stiffness Models}

A stiffness model of force as a function of displacement was fit with a polynomial expression up to the 3rd degree. A nonlinear regression was performed, with the mean final displacement regressed over reference force level in the tension task during evaluation for each haptic training condition. The polynomial components were tested for significance using an F-test. A component was deemed to have a significant effect on the model if its corresponding p-value was less than 0.05.

\begin{figure*}[!t]
\centering
\includegraphics[width=\textwidth]{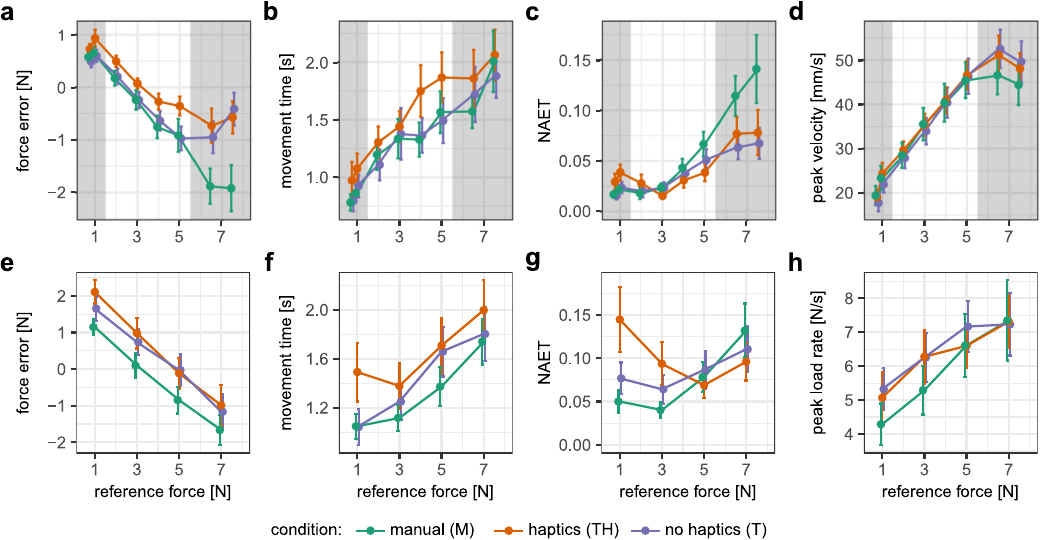}
\caption{(a-d) Mean performance metrics vs. reference force for the tension task. Grey regions denote the range of forces outside those that the participants trained on. (e-h) Mean performance metrics vs. reference force for the palpation task. NAET stands for the Normalized Absolute Error-Time. Error bars denote 95\% confidence intervals.}
\label{results_plots}
\end{figure*}

\section{Results}

\begingroup
\setlength{\thickmuskip}{0.5\thickmuskip}

\subsection{Training Block}
The learning curves for each haptic training condition are summarized as plots of the mean absolute force error against trial number for each reference force level (Fig.\,\ref{Learning curves}). The results indicate that there was a decrease in mean force error over the first 10 trials for all haptic training conditions. Beyond the 10th trial, the mean absolute force error for each condition and force level became more consistent. This suggests that participants had learned to replicate their target reference forces. Towards the end of each set of 30 trials, there was a noticeable increase in force error for several consecutive trials. The increase in force error was more pronounced in the T and TH conditions, for the 4.5\,N and 5.5\,N force levels. This could have resulted from greater mental fatigue due to participants performing teleoperation, which might be more cognitively demanding than manual operation.

\subsection{Evaluation Block}

\subsubsection{Tension Task}
In the tension task, force error values were positive for low reference forces, and became negative as reference force levels increased. This indicates that there was consistent overestimation at lower reference force levels followed by consistent underestimation at higher force levels (Fig.\,\ref{results_plots}a). Hypothesis testing found a significant effect of the haptic training condition ($p=0.037$). There was a significant interaction between reference force level and haptic training condition ($p<0.001$). Post-hoc testing found a significant difference between the M condition and both the teleoperated conditions at 6.5\,N \mbox{(M-TH:\,$p<0.001$; M-T:\,$p<0.001$)} and \mbox{7.5\,N (M-TH:\,$p<0.001$; M-T:\,$p<0.001$)}. These were the two force levels that were above the range of forces that the participants had trained on. At these force levels, the M group was underestimating the forces compared to the teleoperated groups (Fig.\,\ref{results_plots}a). Additionally, there was a significant difference in force error between the M and TH condition at the 4\,N ($p=0.019$) and 5\,N ($p=0.001$). There was also significant interaction between reference force level and experiment progression ($p=0.034$), with higher reference force levels showing less decrease in force error as the experiment progressed compared to lower reference force levels.


Movement time increased as reference force increased, with no significant difference found between conditions (Fig.\,\ref{results_plots}b). Movement time decreased with experiment progression, with the effect being statistically significant ($p=0.007$). There was also a significant interaction between reference force level and experiment progression ($p=0.006$), with higher reference force levels showing less decrease in movement time as the experiment progressed compared to lower reference force levels.

NAET increased as the reference force level increased (Fig.\,\ref{results_plots}c) with the haptic training condition showing a significant effect ($p=0.009$). There was also a significant interaction between the reference force level and the haptic training condition ($p<0.001$). Post-hoc testing found a significant difference between the the M condition and the teleoperated conditions at \mbox{6.5\,N (M-TH:\,$p<0.001$; M-T:\,$p=0.008$)} and \mbox{7.5N\, (M-TH:\,$p<0.001$; M-T:\,$p=0.002$)}. This indicated that at those force levels, the M group had worse speed-accuracy functions compared to the teleoperated groups. 

Up to the 6.5\,N force level, the peak velocity increased as the reference force increased. The decrease at 7.5\,N is possibly due to this force being closer to the limit that participants were warned about so that they were likely more cautious as a result. There was an overall significant interaction between the reference force level and the haptic training condition ($p=0.004$). Post-hoc testing did not indicate any significance between haptic training condition groups at any force level. For a detailed description of the statistical results, readers are directed to the accompanying supplementary materials. 

\subsubsection{Palpation Task}

Except for NAET, the same trends with respect to reference force level were reflected in the palpation task as in the tension task (Fig.\,\ref{results_plots}e, f, and h). With respect to the haptic training condition, the force error showed a distinct separation between the teleoperated groups and the M condition, with the M condition showing a consistent underestimation across all force levels compared to the former (Fig.\,\ref{results_plots}e). Hypothesis testing revealed significant interaction between reference force level and experiment progression ($p=0.034$), with higher reference force levels showing less increase in force error as the experiment progressed compared to lower reference force levels. Movement time decreased as the experiment progressed, with the effect being statistically significant ($p=0.043$).

For NAET, there was a significant effect of condition ($p<0.001$) and a significant interaction between the reference force level and the haptic training condition \mbox{($p<0.001$)} to the TH condition at \mbox{1\,N (M-TH:\,$p<0.001$; T-TH:\,$p=0.029$)}. This indicates that the TH condition had worse speed-accuracy functions at the first force levels compared to both the M and T conditions. Additionally, NAET was significantly lower for the M condition compared to the TH condition at \mbox{3\,N ($p=0.006$)}, indicating that the TH condition still had a worse speed-accuracy function compared to M, while the difference was eliminated for the T condition.

For peak load rate, there was a significant interaction between the reference force level and the haptic training condition ($p<0.025$). However, post-hoc testing revealed no significant differences between conditions at every force level. For a detailed description of the statistical results, readers are directed to the accompanying supplementary materials. 

\subsection{Estimated Stiffness Models}

The polynomial models that were fit to the mean displacement over each reference force level had statistically significant quadratic terms for the M and TH conditions and a statistically significant cubic term for the T condition (Fig.\,\ref{model_plot}). For the latter, the fit indicates an initially quadratic relationship of force as a function of displacement that becomes approximately linear as reference force level increases. The shapes of the curves also highlight the observation from Fig.\,\ref{results_plots}a that there is initial overestimation and later underestimation for all haptic training conditions as the force level increases.

\endgroup

\section{Discussion}


Our results can be interpreted in three regimes that were present in the experiment: The performance in the learned task (tension) over learned forces, performance in the learned task at novel forces, and performance in a novel task (palpation). 

For the learned task, we found that learned dynamics from training influence force estimation abilities in evaluation. During performance in the learned task over learned forces, there was no significant difference between groups and no evidence of the use of a haptically informed, learned stiffness model. However, the systematic underestimation of force at low reference force levels and overestimation at high reference force levels suggest the use of a force estimation heuristic that is reliant on arm proprioception \cite{Fuentes2010}. Alternatively, this observed trend could have been due to a bias towards the mean of movements learned during training.

In the learned task at novel high force levels, there was similar force estimation performance by the participants in the T and TH groups and a significant underestimation of force by the participants in the M group. This result contrasts with our expectation that the haptics groups should perform similarly, given our hypothesis that haptic experience leads to better performance. Thus, we attribute the results to either a lack of adaptation to correct for biases in proprioception, or to unfamiliarity with the new dynamics found in teleoperation, leading to stronger bias towards the mean movement learned during training compared to the T and TH groups. 

In the novel task, there was a small effect in performance as measured by speed-accuracy at low reference force levels for the M condition compared to the TH condition. This suggests that participants could access a haptically informed stiffness model of the sample, which was combined with models of viscoelastic materials from daily experience, to extrapolate the sample stiffness in compression and thus improve their speed-accuracy.

Additionally, in the tension task, we found that prior haptic training experience does result in participants better capturing the non-linearity of a stiffness estimate at higher force levels.  

\subsection{Performance in Learned Task and Forces}

The similarities in force error (Fig.\,\ref{results_plots}a) and NAET (Fig.\,\ref{results_plots}d) between all conditions below 5.5\,N suggest that when movements and forces are similar to those that were previously learned, there is no use of a haptically informed stiffness model. Instead, the consistent trend across all groups, of underestimation at low reference force levels and overestimation at high reference force levels, supports the use of a displacement-force heuristic. 

One explanation is human proprioception bias. Fuentes and Bastian have shown that there is a bias in the estimate of one's arm position towards more extreme positions such that, when the elbow is extended to an obtuse angle, it is perceived to be at a more obtuse angle than it actually is \cite{Fuentes2010}. Likewise, if the elbow is at an acute angle, it is perceived to be at a more acute angle than it actually is. At high force levels and corresponding high displacements, this belief that the arm is more extended than it really is leads to systematic overestimation of imparted sample displacement and results in force undershoot. At low force levels and low displacements, the effect is the opposite, and the belief that the arm is less extended than it really is leads to systematic underestimation of imparted sample displacement and results in force overshoot.

Another possible explanation for the observed trend is that the movements made during evaluation are biased towards the mean of the movements learned during training. This has been documented in reaching experiments where people experienced linearly increasing forces, but planned movements based on the average of the previously experiences forces as opposed to predicting the increase \cite{Mawase2012}. Under such assumptions, the observed force error would be smallest at the mean reference force of 3.5\,N experienced during training. In Fig.\,\ref{results_plots}a, we see that the point at which the force error crosses zero is close to 3\,N, thus lending evidence to this possibility. While the participants in the M and TH conditions learned to perform the task while experiencing environmental forces, those in the T condition did not. Because the experiments by Mawase and Karniel only investigated the effect of prior experience on movements under changing task dynamics \cite{Mawase2012}, the identical trend seen in those under the T condition remains unaddressed by this possible explanation. 

\subsection{Generalization in Learned Task to Novel Forces}

The high reliance on proprioception for displacement judgments is consistent with the significantly greater force undershoot by the M group compared to the T and TH groups at the reference force levels outside the training range (Fig.\,\ref{results_plots}a). During training, the teleoperated group (T and TH) operated under a motion scaling ratio of 2:1, which is identical to that in the evaluation block. Meanwhile, the M group trained under a different motion scaling of 1:1. Because the M group never had to estimate force at the large arm displacements encountered by the T and TH groups in training, participants in the M group might not have compensated for their arm position overestimation bias when encountering force-displacement ranges outside what they were exposed to. The learned adaptation to the different dynamics of teleoperation has been previously documented by Nisky et al. in a study of reaches by novices and experienced surgeons \cite{Nisky2014}.

As discussed by Mawase and Karniel, the effect of a bias toward the mean of past experiences of force might be influenced by the familiarity of altered dynamics of the task \cite{Mawase2012}. Hence, under more well interpreted dynamics, such as object lifting as opposed to viscous force fields, this effect is eliminated \cite{Mawase2010}. In our experiment, the M group experienced a change in motion scaling ratio, and a switch from manual manipulation to teleoperation, between training and evaluation phases, which could result in greater unfamiliarity with the altered dynamics, thus leading to a stronger effect of the bias towards the mean at previously unseen levels of force. 

The larger force underestimation of the M group, compared to the T and TH groups, occurred at force levels relevant to knot-tying and suturing \cite{Toledo1999}. Given that a proprioception-force heuristic might have be used to perform the tensioning task, underestimation of forces could mean that poor adaptation to task dynamics, like that seen in the M group, could lead to poor performance of the tension-critical portions of suturing and knot-tying maneuvers in RMIS. 

\begin{figure}[!t]
\centering
\includegraphics[width=1\linewidth]{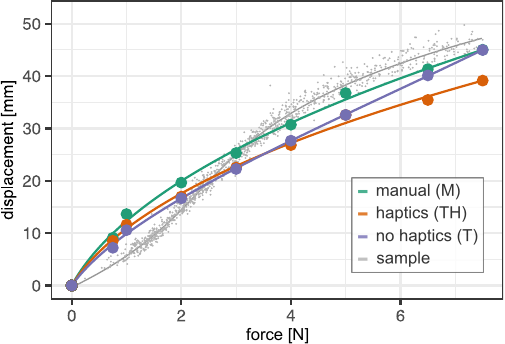}
\caption{Polynomial models of force as a function of displacement compared to the actual silicone sample stiffness curve (grey). The mean displacement was regressed over reference force level for the three haptic training conditions. Large dots are data points representing the mean displacement at a reference force. Small dots are data points collected from the actual silicone sample.}
\label{model_plot}
\end{figure}

\subsection{Generalization to Novel Task}

The required movement direction in the tension task is horizontal, which is different from the vertical direction of movement required in the palpation task. This makes it unlikely that the same failure to adapt to proprioceptive bias seen in the tension task would have contributed to a consistent bias towards lower force estimates, and a resultant lower NAET for the M group compared to the teleoperated groups. Likewise, there was no prior experience of the dynamics of the palpation task, and hence the effect of any bias towards a mean over prior experience would be mitigated. Instead, the small separation of NAET between the M and TH groups at low force levels (Fig.\,\ref{results_plots}g) provides some evidence that participants belonging to M group were able to leverage past interactions with viscoelastic materials in compression during daily living, to improve force estimation performance. 

Through probabilistic mechanisms described by Di Luca and Ernst \cite{DiLuca2014}, the sample stiffness experienced by the participants over the training phase would have been integrated with an estimate of stiffness derived from viscoelastic objects they previously experienced over the course of their daily life. These prior models of stiffness would naturally contain a compression component in addition to a tension component. Since stiffness is primarily a relationship between force and displacement, the stiffness model would be correct only if the scale of interaction was similar to that of daily living. This can only occur in the M training condition, and thus the access of such a haptically informed stiffness model offers an explanation for the lower NAET of the M group compared to the TH group. Because force information was still available through color change, the better performance of the T group compared to the TH group in NAET is possibly explained by the former having developed better accuracy in interpreting the color change as a force cue during training. 

The better speed-accuracy function of the  M group compared to the TH group occurred at force levels relevant for palpation in minimally invasive surgery \cite{Toledo1999}. However, they are below the average forces reported for dissection, another maneuver that can require compression forces \cite{Wagner2007}\cite{Toledo1999}. Thus, the benefits of the better speed-accuracy function seen for the M group might imply better performance in RMIS only for palpation. 

\subsection{Estimates of Stiffness}
While force estimation performance of participants during evaluation were differentiated by the task dynamics experienced in training, the resultant estimated stiffness models they formed for the tension task in evaluation were instead differentiated by the availability of haptic feedback in training. The shape of the estimated force-displacement curve for the T condition in Fig.\,\ref{model_plot} suggests that participants were performing a more linear extrapolation for the forces outside the training range compared to the M and TH groups. Thus, while no group correctly estimated the non-linear stiffness of the silicone throughout the entirety of the tested force range, the haptics groups did capture the upward concavity of the force-displacement curve that was characteristic of the sample at the higher force range. This was reflected analytically in the cubic terms of the regression fits for the M and TH groups, which were not found to be significant, unlike that of the T group. This is consistent with the results of Wu and Klatzy, who provide evidence of a model of stiffness that recursively updates throughout an interaction \cite{Wu2018}. They also found that the final estimate of stiffness is highly weighted towards that which was experienced towards the end of the interaction. In our experiment, the haptic training conditions (M and TH) provided more continuous force information compared to the post-trial error feedback available to the T group. This would have allowed the participants in the haptic training conditions to capture more of the non-linearities of the sample stiffness and consequently weight the upward concave stiffness experienced during the latter part of the interaction more highly, thus arriving at the quadratic stiffness estimate.  

\subsection{Future Work}

In this experiment, participants were not given instructions to make fast movements, which could have contributed to failure to learn a prior of stiffness during the tensioning task. Instructions that shift attention away from internal movement control contribute to movement automatization \cite{Shea1999}\cite{Kal2013}, or implicit learning \cite{Kal2018}, which promotes better retention and generalization \cite{WulfReview2013}. Due to better cognitive efficiency, implicitly learned movements result in better performance, and are thus rewarded more under time constraints, compared to explicitly learned movements \cite{Masters2008}. Hence, the provision of an instruction to make fast movements could have promoted implicit learning of a cognitively efficient stiffness model as opposed to an explicit heuristics-based strategy to perform the tasks. Future work should investigate if the promotion of fast movements during the training phase can encourage the development and use of stiffness models to improve performance and generalization of force estimation in the learned task.   

While the results of both evaluation tasks suggest that the amount of motion scaling during training is likely to have an effect on force estimation ability, only one ratio of scaling was investigated in this study. To further investigate the effect of scaling on force estimation in RMIS, future experiments should train participants at different ratios of motion scaling.

Because of its simplicity and short timescale, the results of this study might not be fully relevant to the experience of clinical RMIS training, which is known to involve tasks of greater complexity and duration. The tension task is simple compared to those in other haptics-related RMIS studies like blunt dissection, which requires making a variety of scraping movements in multiple directions \cite{Wagner2007}, needle driving, which requires grasping a needle and driving it along a curved path through soft material \cite{Bahar2020}, and even peg transfer, which involves grasping and lifting rubber samples from a peg with minimal damage and high positional accuracy \cite{King2009}. The above tasks engage implicit mechanisms of force estimation to achieve higher-order goals, and are similar to the critical points at which attendings gauge trainees’ force sensitivity in rubrics such as GEARs. Our simple task on the other hand, requires participants to reason directly in terms of forces and is likely to engage more explicit mechanisms of force estimation. Thus, participants might not have been required to develop a prior representation of stiffness of the material to achieve satisfactory performance, instead successfully relying on a heuristic mapping of force to displacement. While there has been evidence to show that explicit and implicit mechanisms can be developed in parallel even for simple tasks such as point-to-point reaching \cite{Taylor2014}, it is also possible that the two are independent \cite{Leib2015}. Thus, it is unclear if our work can generalize to more complex tasks like those found in RMIS. The short duration of the study also might not have afforded enough time for participants to learn a prior representation of stiffness, as it typically takes expert surgeons a long period of time to develop good tissue-handling skills with a surgical robot. This slower rate of implicit learning has been documented in prior work \cite{Taylor2014}. Furthermore, there is evidence suggesting that skill retention is less for explicit learning compared to implicit learning \cite{Benson2011}. 

Our experiment sought to mitigate these limitations by focusing on tasks with simple dynamic models, learned under conditions that reduce reliance on external cues. Our findings suggest that prior haptic experience potentially helps with the generalization of force estimation to simple, unseen tasks. Future work is needed to build upon our basic result, and investigate if haptically informed models of stiffness can be learned in more complex and realistic tasks, likely over longer periods of time. Additionally, the use of a more complex, clinically relevant task would facilitate baseline comparisons with expert surgeons, and thus offer greater insight into how meaningful any effect of haptic training would be in developing force estimation skill in RMIS.


\section{Conclusion}

This study tested the effects of prior haptic training in two modalities, manual manipulation and teleoperation, on force estimation performance during the teleoperation of a minimally invasive surgical robot in a learned task (tension) and a novel task (palpation) against a control condition of teleoperation without haptics. In both the learned and novel tasks, our results pointed to a strong influence of task dynamics, specifically the amount motion scaling, during training, on the ability to estimate force with the aid of heuristics and learned models respectively. Firstly, a difference in motion scaling could lead to poor compensation of proprioceptive bias in movements, or to a bias towards the mean of past movements as seen in the learned task. Secondly, in terms of generalization to a novel task, a mismatch of motion scaling could affect the ability to access prior models developed under one scale while estimating forces in another. 

In this work, we studied the effect of prior haptic experience on force estimation in teleoperation, for controlled one-dimensional tension, and compression tasks. While these tasks allowed for straightforward quantitative analysis, they required explicit force estimation, which is unlike the implicit force estimation used to achieve higher-order goals for more complex tasks such as those in RMIS. Therefore it is unclear if our results will generalize outside of the scope of our experiment. Should future work verify our findings in more complex scenarios, it would potentially offer new ways in which haptics can contribute to skills development in RMIS.

\section*{Acknowledgment}

The authors thank I. Farlie, Y. A. Oquendo, and M. R. Lee for their assistance and advice with this project.

\ifCLASSOPTIONcaptionsoff
  \newpage
\fi



%

\bibliographystyle{IEEEtran}
\bibliography{IEEEabrv,references}{}

\end{document}